\documentclass[sigconf]{acmart}
\AtBeginDocument{%
  \providecommand\BibTeX{{%
    \normalfont B\kern-0.5em{\scshape i\kern-0.25em b}\kern-0.8em\TeX}}}

\setcopyright{none}
\usepackage{multirow}
\usepackage{subcaption}
\begin{document}

%%
%% The "title" command has an optional parameter,
%% allowing the author to define a "short title" to be used in page headers.
\title{AnomalySD: Few-Shot Multi-Class Anomaly Detection with Stable Diffusion Model}

%%
%% The "author" command and its associated commands are used to define
%% the authors and their affiliations.
%% Of note is the shared affiliation of the first two authors, and the
%% "authornote" and "authornotemark" commands
%% used to denote shared contribution to the research.

\author{Zhenyu Yan}
\affiliation{%
  \institution{Sun Yat-Sen University}
  % \streetaddress{1 Th{\o}rv{\"a}ld Circle}
  \city{Guangdong}
  \country{China}}
\email{yanzhy9@mail2.sysu.edu.cn}

\author{Qingqing Fang}
\affiliation{%
  \institution{Sun Yat-Sen University}
  % \streetaddress{1 Th{\o}rv{\"a}ld Circle}
  \city{Guangdong}
  \country{China}}
\email{fangqq3@mail2.sysu.edu.cn}

\author{Wenxi Lv}
\affiliation{%
  \institution{Sun Yat-Sen University}
  % \streetaddress{1 Th{\o}rv{\"a}ld Circle}
  \city{Guangdong}
  \country{China}}
\email{lvwx8@mail2.sysu.edu.cn}

\author{Qinliang Su}
\affiliation{%
  \institution{Sun Yat-Sen University}
  % \streetaddress{1 Th{\o}rv{\"a}ld Circle}
  \city{Guangdong}
  \country{China}}
\email{suqliang@mail.sysu.edu.cn}
%%
%% By default, the full list of authors will be used in the page
%% headers. Often, this list is too long, and will overlap
%% other information printed in the page headers. This command allows
%% the author to define a more concise list
%% of authors' names for this purpose.
\renewcommand{\shortauthors}{Zhenyu Yan and Qingqing Fang and Wenxi Lv and Qinliang Su, et al.}

%%
%% The abstract is a short summary of the work to be presented in the
%% article.
\begin{abstract}
Anomaly detection is a critical task in industrial manufacturing, aiming to identify defective parts of products. Most industrial anomaly detection methods assume the availability of sufficient normal data for training. This assumption may not hold true due to the cost of labeling or data privacy policies. Additionally, mainstream methods require training bespoke models for different objects, which incurs heavy costs and lacks flexibility in practice.
To address these issues, we seek help from Stable Diffusion (SD) model due to its capability of zero/few-shot inpainting, which can be leveraged to inpaint anomalous regions as normal. In this paper, a few-shot multi-class anomaly detection framework that adopts Stable Diffusion model is proposed, named AnomalySD. To adapt SD to anomaly detection task, we design different hierarchical text descriptions and the foreground mask mechanism for fine-tuning SD. In the inference stage, to accurately mask anomalous regions for inpainting, we propose multi-scale mask strategy and prototype-guided mask strategy to handle diverse anomalous regions. Hierarchical text prompts are also utilized to guide the process of inpainting in the inference stage. The anomaly score is estimated based on inpainting result of all masks.
Extensive experiments on the MVTec-AD and VisA datasets demonstrate the superiority of our approach. We achieved anomaly classification and segmentation results of 93.6\%/94.8\% AUROC on the MVTec-AD dataset and 86.1\%/96.5\% AUROC on the VisA dataset under multi-class and one-shot settings.
\end{abstract}

\settopmatter{printacmref=false}
\maketitle

\section{Introduction}
Anomaly detection is a critical computer vision task in industrial inspection automation. It aims to classify and localize the defects in industrial products, specifically to predict whether an image or pixel is normal or abnormal~\cite{mvtec}. Due to the scarcity of anomalies, existing methods typically assume that it is possible to collect training data from only normal samples in the target domains. Therefore, existing mainstream anomaly detection methods mainly follow the unsupervised learning manner and can be divided into two paradigms, i.e.,reconstruction-based~\cite{smai,scadn,riad} and embedding-based methods~\cite{patchcore,spade,padim}. Reconstruction-based methods mainly use generative models such as Autoencoders (AE) ~\cite{ae} or Generative Adversarial Networks (GAN) ~\cite{gan} to learn to reconstruct normal images and assume large reconstruction errors when reconstructing anomalies. While embedding-based methods aim to learn an embedding neural network to capture embeddings of normal patterns and compress them into a compact embedding space~\cite{patchcore,padim}.

\begin{figure}[t!]
    \centering
    \begin{subfigure}[t]{0.73\columnwidth}
        \centering
        \includegraphics[height=1.15in]{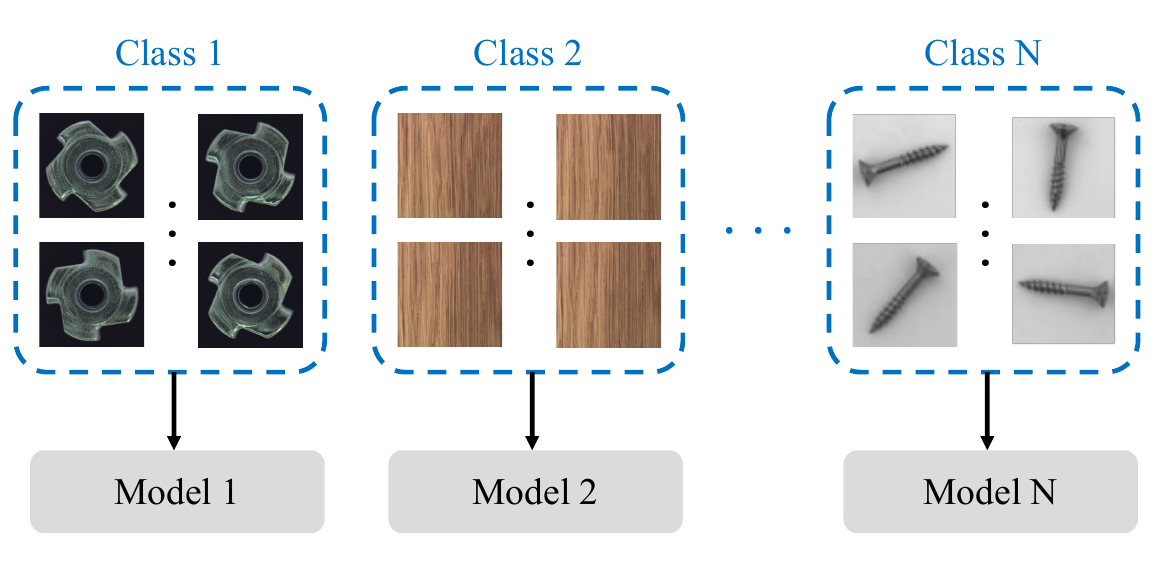}
        \caption{Unsupervised Anomaly Detection (Vanilla)}
    \end{subfigure}%
    \hfill
    \begin{subfigure}[t]{0.23\columnwidth}
        \centering
        \includegraphics[height=1.15in]{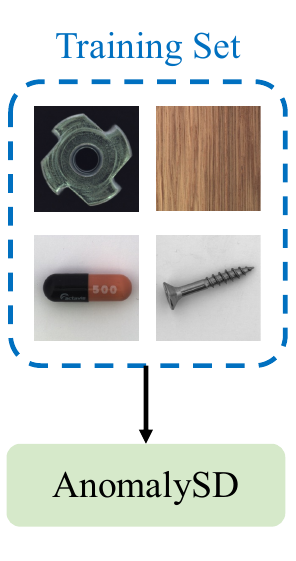}
        \caption{Ours}
    \end{subfigure}
    \vspace{-0.2cm}
    \caption{Difference between (a) vanilla unsupervised anomaly detection and our (b) few-shot anomaly detection under multi-class setting.}
    \label{fig:one-for-one}
    \vspace{-0.4cm}
\end{figure}

%Most of main-stream method train bespoke model for every category and required hundreds normal data
Although the astonishing performance achieved by previous methods, most of them assume that there are hundreds of normal images available for training. But in the real world, this assumption cannot always be fulfilled. It is likely that there are only a few normal data available due to the high cost of labeling or data privacy policies, which is called the few-normal-shots setting. Under such a scenario, performance degeneration of traditional anomaly detection methods is observed as they require abundant normal data to well capture the normal pattern~\cite{tdg}. Parallel to this, the bulk of existing works follow the one-for-one paradigm as shown in Figure \ref{fig:one-for-one}(a), which needs to train a bespoke model for each category. This paradigm results in heavy computational and memory costs and more resources are required to store different models. Moreover, the one-for-one paradigm is not flexible for real-world applications. In practice, different components may be produced on the same industrial assembly line, which needs to switch different models to detect different anomalies or deploy multiple networks with different weights but the same architecture. The additional expenses caused by the lack of flexibility are completely prodigal.

%current multi-class study --> existing issue: still require hundreds of normal data --> current few-show --> existing issue: still need to train bespoke model and leverage inappropriate datasets
The issues mentioned above have been already noticed in the most current study. 
%few-show method 
Given only a few normal data, enlarging the training dataset~\cite{graphcore, regad} and patch distribution modeling~\cite{tdg, diffnet} are promising approaches. 
GraphCore~\cite{graphcore} directly enlarges the normal feature bank through data augmentation and trains a graph neural network on it to figure out anomalies. RegAD~\cite{regad} is based on contrastive learning to learn the matching mechanism, which is used to detect anomalies, on an additional dataset. While TDG~\cite{tdg} detects anomalies by leveraging a hierarchical generative model that captures the multi-scale patch distribution of each support image. DiffNet~\cite{diffnet} uses normalizing flow to estimate the density of features extracted by pre-trained networks.
But existing few-shots anomaly detection methods do not take the multi-class setting into consideration.
%multi-class method
Recently, reconstruction-based methods ~\cite{unified,oneforall, diad} show the potentiality of handling the multi-class anomaly detection task. To train a unified model having the capability of reconstructing normal images of different categories, ``identical shortcut" is the key issue to solve. The ``identical shortcut" issue~\cite{unified} is caused by the complexity of multi-class normal distribution resulting in both normal and anomalous samples can be well recovered, and hence fail to detect anomalies.
UniAD~\cite{unified} adopts Transformer architecture and proposes neighbor masked attention mechanism, which prevents the image token from seeing itself and its similar neighbor, to alleviate the issue. Instead of directly reconstructing the whole normal data, PMAD~\cite{oneforall} utilizes Masked Autoencoders to predict the masked image token based on unmasked tokens. Due to the powerful capability of image generation, diffusion model~\cite{ddpm} is also applied to the multi-class anomaly detection task. DiAD~\cite{diad} is proposed to leverage the latent diffusion model to capture normal patterns of different categories and calculate the reconstruction error in feature space, achieving excellent performance. However, all of these multi-class methods require using a large number of normal data to capture the normal patterns across different categories. 
%Training a unified anomaly detection model that could be applied to many classes of data with only few-short normal data available is still a challenging problem. 
Recently, WinCLIP~\cite{winclip} has shown the potentials of using pre-trained vision-language models (VLM) for zero/few-shot multi-class anomaly detection. In WinCLIP, it first designs many textual descriptions to characterize the anomalies from different aspects and then uses a vision-language-aligned representation learning model like CLIP~\cite{clip} to assess the conformity between the textual descriptions of anomalies and image patches, which is then used to decide the normality of a patch. However, this representation-matching-based method heavily relies on the quality of textual descriptions for anomalies. But as the anomalies could appear in countless forms in practice, the requirement of high-quality textual descriptions for every possible type of anomalies could hinder its applicability in complex scenarios.

Instead of using CLIP to assess the conformity between the textual anomaly descriptions and image patches, in this paper, we propose to address the problem based on a totally different approach by exploiting the powerful inpainting capability of pre-trained Stable Diffusion (SD) model~\cite{ldm}. Our idea is that we first manage to obtain some masks that can largely cover the anomalous areas in images and then use the SD to inpaint the masked anomalous areas into normal ones with the unmasked normal pixels. By comparing the original image and the inpainted image, ideally the anomalous region can be localized. Since SD is pre-trained for general image inpainting tasks, to make it more suitable to the anomaly detection task, we first propose to a method to finetune the model, boosting its capability in inpainting the anomalous areas into normal ones. To this end, we propose a foreground masking mechanism to randomly mask some areas in an image and then encourage the SD to recover the masked area by finetuning. To further enhance the model's recovering-to-normal ability, textual prompts are further designed and used to guide the model to inpaint the anomalous area into normal one during the denoising process. 
To use the finetuned SD for anomaly detection and localization, our idea is to propose many possible anomaly areas and then use the finetuned SD to inpaint them into normal ones, with the final anomaly scores estimated from all of these inpainting results. To obtain masks that can largely cover the true anomalies, due to the various forms of anomalies, a multi-scale mask strategy that generates different sizes of rectangular masks is employed. Moreover, to break the constraint of rectangular shape in mask, a prototype-guided mask strategy is further proposed, which is able to generate mask with more flexible shapes and sizes. Experimental results on the widely used industrial anomaly datasets MVTec and VisA show that our model achieves highly competitive performance comparing with existing methods under the few-shot multi-class setting. 

\section{Related Work}
\paragraph{\textbf{Anomaly Detection}}
The mainstream anomaly detection methods can be divided into two trends: embbeding-based methods, reconstruction-based methods. Embedding-based methods primarily detect anomalous samples in the feature space. Most of these methods utilize networks pre-trained on ImageNet ~\cite{imagenet} for feature extraction and then calculate the anomaly score by measuring the distance between anomalous and normal samples in the feature space ~\cite{padim, patchcore, spade}. While some embbeding-based methods employ knowledge distillation to detect anomalies based on the differences between teacher and student networks ~\cite{uninformed, distillation}. 
Reconstruction-based methods primarily aim to train the model to learn the distribution of patterns in normal samples. AE-based methods ~\cite{memae, robustae} and inpainting methods ~\cite{smai,scadn,riad} are both based on the assumption that the model can effectively reconstruct normal images but fail with anomalous images, resulting in large reconstruction errors. 
In the case of VAE-based generative models ~\cite{iterative, towards} learn the distribution of normal in the latent space. Anomaly estimation is carried out by assessing the log-likelihood gap between distributions. For GAN-based generative models ~\cite{anogan, f-anogan, skipgan, ocrgan, ganomaly, dividegan}, the discriminator compares the dissimilarity between test images and images randomly generated by the generator as a criterion for anomaly measurement.

\paragraph{\textbf{Few-Shot Anomaly Detection}} 
Few-shot anomaly detection is developed for the situation where only a few normal data are available. 
TDG~\cite{tdg} proposed to leverage a hierarchical generative model that learns the multi-scale patch distribution of each support image.
DiffNet~\cite{diffnet} normalizing flow to estimate the density of features extracted by pre-trained networks.
To compensate for the lack of training data, RegAD~\cite{regad} introduced additional datasets to learn the matching mechanism through contrastive learning and then detect anomalies by different matching behaviors of samples.
Although these methods can handle the few-shot anomaly detection task, none of them take the multi-class setting into consideration. Recently, WinCLIP~\cite{winclip} revealed the power of pre-trained vision-language model in few-shot anomaly detection task. It divides the image into multi-scale patches and utilizes CLIP~\cite{clip} to calculate the distance between patches and the designed description of normality and anomaly as the anomaly score. Thanks to the generalization ability of CLIP~\cite{clip}, WinCLIP also has the ability to handle multi-class setting.

\paragraph{\textbf{Multi-Class Anomaly Detection}}
Multi-class anomaly detection methods aim to develop a unified model to detect anomalies of different categories to save computational resources. Most of them are based on the reconstruction-based paradigm~\cite{unified, oneforall, diad}. UniAD~\cite{unified} adopted Transformer architecture to reconstruct image tokens extracted by pre-trained networks and proposed neighbor masked attention mechanism, which prevents the image token from seeing itself and its similar neighbor, to alleviate the ``identical shortcut". PMAD~\cite{oneforall} deems that the objective of reconstruction error results in more severe ``identical shortcut" issue, they use Masked Autoencoders architecture to predict the masked image token by unmasked image tokens. The uncertainty of predicted tokens measured by cross-entropy is used as anomaly score. Due to the diffusion model's powerful capability of image generation, DiAD~\cite{diad} adopted the latent diffusion model to reconstruct normal images and calculate the anomaly score in feature space.
However, all of them require plenty of normal images to capture normal patterns while our method needs only a few normal images for fine-tuning. 

\paragraph{\textbf{Anomaly Detection Based on Diffusion model}}
Diffusion models ~\cite{ddpm, ddim} have garnered attention due to their powerful generative capabilities, aiming to train the model to predict the amount of random noise added to the data and subsequently denoising and reconstructing the data. Stable Diffusion Model (SD) ~\cite{ldm} introduces condition through cross-attention, guiding the noise transfer in the generation phase towards the desired data distribution. In the field of anomaly detection, AnoDDPM ~\cite{anoddpm} has made initial attempts to apply diffusion models in reconstructing medical lesions within the brain. DiffusionAD ~\cite{diffusionad} comprises a reconstruction network and a segmentation subnetwork, using minimal noise to guide the reconstruction network in denoising and reconstructing synthetic anomalies into normal images. AnomalyDiffusion ~\cite{diffusiongeneration} leverages the potent generative capabilities of Diffusion to learn and generate synthetic samples from a small set of test anomalies. However, current Diffusion-based anomaly detection methods primarily focus on denoising without sufficient exploration of how to systematically and controllably reconstruct anomalies into normal samples.

\section{Preliminaries}
\subsection{Denoising Diffusion Probabilistic Model} 
\label{sec:ddpm}
The Denoising Diffusion Probabilistic Model (DDPM)~\cite{ddpm} is a type of generative model inspired by the data diffusion process. DDPM consists of a forward diffusion process and a reverse denoising process. In the forward diffusion process, at every time step, we add a small noise into the data as $x_t = \sqrt{1-\beta_t}x_{t-1} + \sqrt{\beta_t} \epsilon_t$, where $x_0$ denotes the original image; $\epsilon_t \sim {\mathcal{N}}(0, I)$ represents the standard Gaussian noise; and $\beta_t$ is used to control the noise strength added at the time step $t$, which is often a very small value from $\in(0,1)$. It can be easily shown that $x_t$ can be directly obtained from $x_0$ as
\begin{equation}
x_t = x_0 \sqrt{\bar{\alpha}_t} + \epsilon_t\sqrt{1-\bar{\alpha}_t},
\label{eq:ddpm_forward}
\end{equation}
where $\bar{\alpha}_t = \prod_{i=1}^{t}\alpha_i$ with $\alpha_i \triangleq (1-\beta_i)$. In DDPM, it seeks to learn the reverse process of the forward diffusion process by learning a model to predict $x_{t-1}$ solely based on $x_t$. It is shown in DDPM that $x_{t-1}$ can be predicted as below
\begin{equation}
x_{t-1} = \frac{1}{\sqrt{\alpha_t}} \left( x_t - \frac{1-\alpha_t}{\sqrt{1- \bar{\alpha}_t}}\epsilon_{\theta}(x_t,t) \right) + \tilde{\beta}_t \eta,
\end{equation}
where $\eta \sim \mathcal{N}(0, I)$ is another Gaussian noise and $\tilde \beta_t = \frac{1-\bar \alpha_{t-1}}{1-\bar \alpha_t}\beta_t$; and
$\epsilon_{\theta}(x_t,t)$ denotes the prediction of noise $\epsilon_t$ in \eqref{eq:ddpm_forward}, which in practice is realized by using a U-Net structured network ~\cite{unet}. The model parameter $\theta$ is trained by minimizing the error between the predicted noise $\epsilon_\theta(x_t, t)$ and the truly added noise $\epsilon_t$
\begin{equation}
\mathcal{L} = \mathbb{E}_{x_0 \sim q(x_0), \epsilon \sim \mathcal{N}(0,I),t \sim [1, T]} \parallel \epsilon - \epsilon_{\theta}(x_t,t) \parallel_2^2,
\end{equation}
where $T$ denotes the total number of steps used in the forward diffusion process.

\subsection{Stable Diffusion Model} 
Based on DDPM, the Stable Diffusion Model(SD)~\cite{ldm} introduces a pre-trained AutoEncoder with the encoder $\mathcal{E}(\cdot)$ to compress image $x$ into latent representation $z = \mathcal{E}(x)$ and the decoder $\mathcal{D}(\cdot)$ to recover the latent features to image $\mathcal{D}(z)$. The adaptation of AutoEncoder helps SD to generate high-resolution images in good quality when the DDPM process is utilized in the latent space by replacing $x$ in Sec \ref{sec:ddpm} with $z$. Besides, SD introduces the condition mechanisms to guide the generation of images by the cross-attention modules in the denoising U-Net. For the condition $y$, SD first encodes it by a pre-trained encoder $\tau(\cdot)$, like CLIP text encoder for text condition. Then, SD introduces the condition $y$ into the $i$-th intermediate layer of U-Net with a cross-attention mechanism
\begin{equation}
attention(Q,K,V) = softmax \left( \frac{QK^T}{\sqrt{d}} \right) \cdot V,
\end{equation}
with $Q=W_Q^{(i)}\cdot \epsilon_{\theta}^{(i)}(z_t), K=W_K^{(i)}\cdot \tau(y), V = W_V^{(i)}\cdot \tau(y)$,
where $\epsilon_{\theta}^{(i)}(z_t)$ represents the flattened output from the intermediate layer $i$ of denoising network $\epsilon_{\theta}(\cdot)$, 
% represent query $Q$ from output $g^{(i)}(z_t)$ of the U-Net intermediate layers $\epsilon_{\theta}$, key $K$ and value $V$ from the condition respectively, 
$W_K^{(i)} \in \mathbb{R}^{d\times d^{(i)}_\epsilon}, W_K^{(i)}, W_V^{(i)} \in \mathbb{R}^{d\times d_\tau}$ are learnable weight parameters. Using the attention mechanisms, conditional information like text prompt  can be introduced to guide the denoising process, leading to the training objective function
\begin{equation}
\label{eq:SD}
\mathcal{L}_{SD}=\mathbb{E}_{\mathcal{E}(x),y,\epsilon \sim \mathcal{N}(0,I), t\sim[1,T]}[\parallel\epsilon-\epsilon_{\theta}(z_t, t, \tau(y)\parallel_2^2].
\end{equation}
In our paper, the conditional information $y$ specifically refers to textual prompts and $\tau(\cdot)$ means the text encoder in CLIP. With the learned denoising network $\epsilon_{\theta}(z_t, t, \tau(y))$ in SD, we first sample a standard Gaussian noise $z_T$ and then feed it into the denoising network. After many steps of denoising, a denoised latent representation $z_0$ is obtained, which is then passed into the pre-trained decoder ${\mathcal{D}}(\cdot)$ to produce the image $x$. 

\begin{figure*}[ht]
% \vskip 0.2in
% \begin{center}
\centering
\includegraphics[width=1.8\columnwidth]{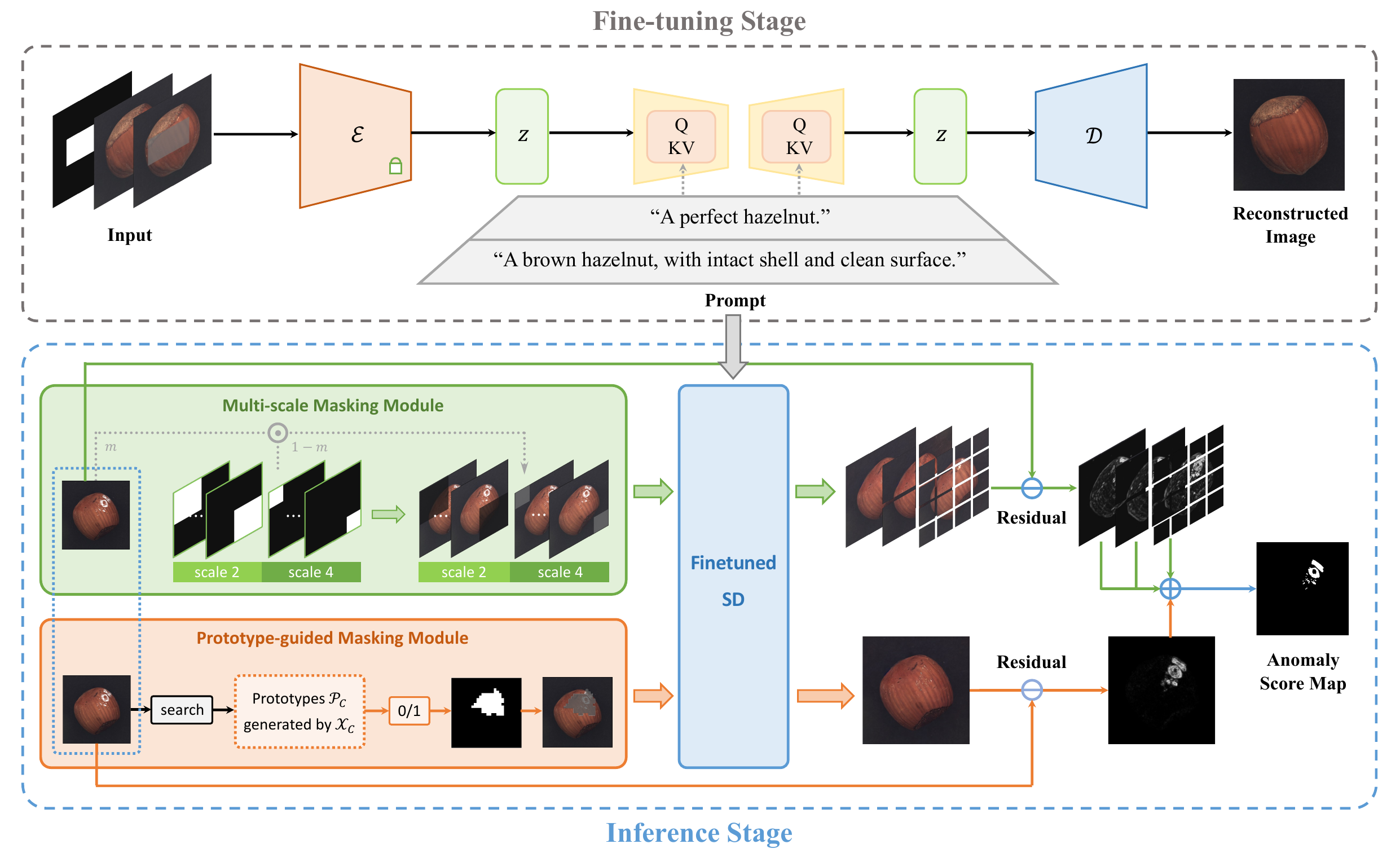}
% \vspace{-0.2cm}
\caption{
The framework structure of AnomalySD is as follows: 
1) Fine-tuning stage: A stable diffusion denoising network is fine-tuned for inpainting on few-shots normal dataset. 
2) Inference stage: Multi-scale Masking Module and prototype-guided Masking Module are utilized to mask potential anomalous regions for inpainting these areas into normal ones. A fusion averaging of error maps from different masks yields the final anomaly score $\mathcal{S}_{map}$.
}
\label{fig:AnomalySD}
% \end{center}
% \vskip -0.2in
\end{figure*}

\section{The Proposed Method}

For the few-shot anomaly detection and localization, we assume the availability of several normal images from multi-classes $\mathcal{C}$: 
\begin{align}
   & \mathcal{X}_N = \{\mathcal{X}_1, \mathcal{X}_2, \cdots, \mathcal{X}_\mathcal{C}\},\nonumber \\
   & \mathcal{X}_c = \{x_{c,1}, \cdots, x_{c,K}\}.
\end{align}
The collected dataset comprises various texture and object images from different classes in which the possible value $c$ can be carpets, bottles, zippers and other categories. The number of images available for each class is limited, with only $K$ shots available, with $K$ very small which can range from 0 to 4. In this paper, our goal is to use the available few-shot normal dataset $\mathcal{X}$ to build a unified model, which can detect and localize anomalies for all categories. The proposed pipeline AnomalySD is shown in Fig.\ref{fig:AnomalySD}. We first introduce how to fine-tune the SD to better adapt it for the anomaly detection task. Then, with the fine-tuned SD, we propose how to leverage the inpainted image to detect and localize anomalies.

\subsection{Adapt SD to Anomaly Detection}
\label{sec:adp SD}
Distinguishing from the SD for image generation tasks, adapting SD for anomaly detection requires to accurately inpaint anomalous areas into normal ones. To achieve this target, we specifically fine-tune the denoising network and decoder of VAE in SD for anomaly detection.
With only few shots of normal samples from multi-classes, to obtain a unified model that can be applied to all categories, we specifically design mask $m$ and prompt $y$ to guide the fine-tuning process of the inpainting pipeline of SD. 

Given an image $x  \in \mathbb{R}^{3 \times H \times W}$ and mask $m \in \{0, 1\}^{H \times W}$, we first encode the original image $x$ and masked image $(1 -m)\odot x$ by the encoder of VAE in SD, get $z = \mathcal{E}(x)$ and $z^{\circ} = \mathcal{E}((1-m)\odot x)$. Through the forward diffusion process, we can get the noisy latent feature of timestamp t by:
\begin{align}
    z_t = z_0 \sqrt{\bar{\alpha}_t} + \epsilon_t\sqrt{1-\bar{\alpha}_t}, \quad \epsilon_t \sim \mathcal{N}(0,I),
\end{align}
where $z_0 = z$ is the feature of timestamp 0. The mask $m$ can also be downsampled to the same size of $z$ and results as $\tilde{m}$. By concatenate the noisy feature $z_t$, masked latent feature $z^{\circ}$ and downsampled mask $\tilde{m}$ in channels, we get input $\tilde{z}_t = [z_t; z^\circ; \tilde{m}]$ for the denosing process.

In the denoising process of the inpainting SD, our aim is to inpaint the masked areas into normal ones according to the unmasked areas. Another text condition prompt $y$ can be used to guide the inpainting of normal patterns. By encoding the prompt $y$ with CLIP text encoder $\tau(\cdot)$, we can get the language features $\tau(y)$. The denoising network $\epsilon_\theta(\cdot)$ is expected to predict the noise according to prompt guidance $\tau(y)$, $\tilde{z}_t$, and timestamp $t$. The denoising network can be optimized by the training object \eqref{eq:diff} to predict noise.
\begin{equation}
\mathcal{L}=\mathbb{E}_{\mathcal{E}(x),y,\epsilon \sim \mathcal{N}(0,I), t}[\parallel\epsilon-\epsilon_{\theta}(\tilde{z}_t, t, \tau(y)\parallel_2^2].
\label{eq:diff}
\end{equation}
With predicted noise, $z_{t-1}$ can be obtained from $\tilde{z}_t$. Repeating the denoising process, finally, we can get the latent feature which is expected to be features of normal images, and we pass it through the decoder in VAE, getting the final inpainted image $\hat{x}$.

\paragraph{\textbf{Training Mask Design}} For accurate anomaly localization, the inpainting framework is expected to focus on object details rather than the background. Therefore, for the object in each image, we generate a foreground mask $m_f \in [0, 1]^{H \times W}$. For the mask used in the fine-tuning process, a randomly generated rectangular mask $m_r$ is first developed, then we can get the mask $m \in \{m_r | IoU(m_r, m_f) > \gamma\}$ where $IoU$ is the intersection over union. The condition filters the mask $m$ fitting the condition that the overlap area between it and the foreground mask $m_f$ surpasses the threshold $\gamma$ for training.

\paragraph{\textbf{Textual Prompt Design}} To help SD inpaint mask areas into normal ones for collected multi-classes, external key descriptions of normal patterns are needed to guide the accurate inpainting process. To achieve the target, different hierarchical descriptions containing abstract descriptions of the whole image and detailed information of parts of the object in images are designed. For each category $c$, text prompts are designed in a pyramid hierarchy, from the most abstract description to the detailed prompts. For the abstract description of category $c$, we design the coarse-grained prompt set $\mathcal{T}_c^1:=$ ``\texttt{A perfect [c].}''. For example, for the category ``\texttt{toothbrush}'', the coarse-grained prompt is ``\texttt{A perfect toothbrush.}''. These abstract prompts help SD to inpainting mask areas into normal ones at the whole semantic level. For detailed descriptions of category $c$, we design the fine-grained prompt set $\mathcal{T}_c^2$, in which prompts describe detailed normal pattern information, like specific description ``\texttt{A toothbrush with intact structure and neatly arranged bristles.}'' of category toothbrush. These detailed prompts guide the SD to focus on inpainting masked areas into normal ones in detail. Combining these two-level language prompts, we get the prompt set $\mathcal{T}_c = \mathcal{T}_c^1 \cup \mathcal{T}_c^2$ for category $c$. In the fine-tuning stage, randomly selected prompt $y \in \mathcal{T}_c$ is chosen for the corresponding image from category $c$, thus different hierarchical information is trained to help inpainting mask areas into normal ones. The detailed designs of prompts for different categories are shown in the Supplementary.

\paragraph{\textbf{Fine-tune Decoder of VAE in SD}} The decoder in VAE used in SD is responsible for converting images from latent space to pixel space. To further guarantee the accurate recovery of normal patterns with high-quality details, we fine-tuned the decoder framework with MSE and LPIPS~\cite{lpips} loss: 
\begin{equation}
    \mathcal{L}_{\mathcal{D}} = \mathcal{L}_{\text{MSE}} + \beta \mathcal{L}_{\text{LPIPS}},
\end{equation}
where the MSE loss is $\mathcal{L}_{\text{MSE}} = \parallel x - \hat{x} \parallel^2_2$ and LPIPS loss is:
\begin{equation}
    \mathcal{L}_{\text{LPIPS}} = \sum_l \frac{1}{H^{(l)}W^{(l)}} \sum_{h, w} \parallel \! w^{(l)} \odot ([\phi^{(l)}(x)]_{h, w} - [\phi^{(l)}(\hat{x})]_{h,w}) \parallel_2^2.
\end{equation}

For the original image $x$ and inpainted image $\hat{x}$, LPIPS loss measures the distance of multi-scale normalized features extracted from different layers of pre-trained AlexNet $\phi(\cdot)$, and the loss of different channels in layer $l$ is weighted by $w^{(l)}$ which has been trained in LPIPS. The $\beta$ for decoder loss is set to $0.1$ during training. 

\subsection{Inpainting Anomalous Areas in Images with Finetuned SD}
\label{sec:infer}
For an image $x$, if there is a given mask $m$ which can mask most parts of the anomalous areas, we can utilize the fine-tuned SD to inpaint the masked area and obtain a inapinted image $\hat{x}$.
In the inference stage of fine-tuned SD, if we use standard Gausian noise $z_T \sim \mathcal{N}(0, I)$ as the starting step of the denoising process, the generated images could show significant variability comparing with the original ones, making them not suitable for anomaly detection and localization.

To address this issue, we control the added noise strength by setting the starting step of the denoising process at  $\tilde{T} = \lambda \cdot T$, where $\lambda \in [0, 1]$. Given the latent feature $z = \mathcal{E}(x)$, we get a noisy latent feature $z_{\tilde{T}}$ according to \eqref{eq:ddpm_forward}. Thus, in the inference stage, the initial input
\begin{equation}
    \tilde{z}_{\tilde{T}} = [z_{\tilde{T}}; z^{\circ}; \tilde{m}]
\end{equation}
is calculated for the denoising network.
After multiple iterations of denoising, the denoised latent feature can be obtained, which we feed into the decoder of VAE to output the final inpainted image $\hat{x}$.

As mentioned before, we assume that the mask $m$ can mask most areas of anomalies. However, the actual anomalous area is not known in real applications, thus the mask used for inpainting needs to be designed carefully so that the anomalous area can be mostly masked and inpainted into normal patterns. To achieve this target, we design two types of masks including multi-scale masks and prototype-guided masks for inpainting in the inference stage.

\paragraph{\textbf{Proposals of Multi-scale Masks}} Anomalies can appear in any area of the image of any size. In order to inpaint anomalous areas of different sizes and positions, we design multi-scale masks to mask the corresponding areas. Specifically, the origin image $x$ can be divided into $k \times k$ patches with $k\in \mathcal{K} = \{1, 2, 4, 8\}$. For every patch $p^{(k)}_{i,j}$ with $i, j \in [1, k]$ in the $k\times k$ squares, we can generate the corresponding mask $M^{(k)}_{i,j}$ to mask the patch and get the masked image. Then, the masked image and mask can be used for inpainting the area, and we can get the inpainted image $\hat{x}^{(k)}_{i,j}$.

\paragraph{\textbf{Proposals of prototype-guided Masks}} The multi-scale masks help inpaint images in different patch scales, but anomalous areas can cross different patches, thus there may still exist anomalous areas in the masked images, reducing the quality of inpainting into normal images. To solve the problem, we propose prototype masks. At the image level, every pixel only has 3 channels that are hard to be used for filtering anomalous pixels, so the pre-trained EfficientNet $\varphi(\cdot)$ is employed for extracting different scale features from different layers. For any image $x$, we can extract its features to get 
\begin{equation}
    \psi(x) = [\varphi^{(2)}(x); \text{Upsample}(\varphi^{(3)}(x))],
\end{equation}
where $\varphi^{(2)}(\cdot)$ and $\varphi^{(3}(\cdot)$ extract features from layers 2 and 3 from $\varphi(\cdot)$ respectively. The feature from layer 3 is upsampled to the same size as that from layer 2 and the two features are combined by the operate concatenation [;] along the channel. For every category $c$, we get the features of few shot normal samples and form its corresponding normal prototype set:
\begin{equation}
    \mathcal{P}_c = \left\{ [\psi(Aug(x))]_{h, w} \mid h \in [1, H'], w \in [1, W'], x \in \mathcal{X}_c \right\},
\end{equation}
in which $H'$ and $W'$ are the height and width of the feature map extract from layer 2, and $Aug(\cdot)$ augment the normal image by rotation to enrich the prototype bank. In the inference stage, for any test sample $x$ from category $c$, we can also get its corresponding $\psi(x)$. Searching the normal prototype bank, an error map $E$ measures the distance to the normal prototype can be obtained as 
\begin{equation}
    [E]_{h,w} = \min_{f\in \mathcal{P}_c} \| [\psi(x)]_{h, w} - f\|^2,
\end{equation}
Due to the limited number of few-shot samples, $E$ is not precise enough to predict anomaly, but it can be used for producing a mask of any size and shape that may contain the anomalous area. Given an appropriately selected threshold $r$, the error map $E$ can be transformed into a binary mask as
\begin{equation}
    M^{(p)} = \text{Upsample}(\mathbb{I}(E > r)),
\end{equation}
where the $(i,j)$-th element of ${\mathbb{I}}( E > r)$ equals to 1 if the condition $[ E]_{ij} > r$ holds, and equals to 0 otherwise; $\text{Upsample}(\cdot)$ means upsampling the binary map ${\mathbb{I}}( E >r)$ to the same size as image, that is, $H\times W$. Now, we can send the image $x$ as well as the prototype-guided mask $M^{(p)}$ into the finetuned SD and obtain the inpainted image $\hat x^{(p)}$. In our experiments, the threshold $r$ is set to a value that can have the summation of variance of elements in the set $\{[E]_{h, w} \mid  [{E}]_{h, w} \geq r\}$ and the variance of element in the set $\{[E]_{h, w} \mid  [{E}]_{h, w} < r\}$ to be smallest, that is, $r = \arg \min_r {Var(\{[E]_{h, w} \mid  [{E}]_{h, w} \geq r\}) + Var(\{[{E}]_{h, w} \mid [{E}]_{h, w} < r\})}$.

\paragraph{\textbf{Aggregation of Hierarchical Prompts}} In the inference stage, to obtain well inpainted normal image, different hierarchical prompts are aggregated together as the condition guidance. Specifically, average prompt representation Avg(\{$\tau(y) | y \in \mathcal{T}_c$\}) is used as the condition prompt guidance for images from category $c$. 

\subsection{Anomaly Score Estimation}
For a test image $x$, we can get its corresponding inpainted image $\hat{x}$ which are assumed to be recovered into normal ones. However, the difference between inpainted and original images is hard to measure at pixel level which only has 3 channels. Considering LPIPS loss used in fine-tuning the decoder of VAE, we also use features extracted from pre-trained AlexNet to measure the difference between $x$ and $\hat{x}$. For the features extracted from layer $l$ of AlexNet, the distance of original and inpainted images in position $(h, w)$ of corresponding feature maps can be calculated as
\begin{equation}
    [D^{(l)}(x, \hat{x})]_{h, w}= \parallel w^{(l)} \odot ([\phi^{(l)}(x)]_{h, w} - [\phi^{(l)}(\hat{x})]_{h,w}) \parallel_2^2.
\end{equation}
We upsample $D_l$ to the image size and add distances from all layers to get the score
\begin{equation}
    D(x, \hat{x}) = \sum_{l} \text{Upsample}(D^{(l)}(x, \hat{x})).
    \label{eq: distance}
\end{equation}

For the multi-scale masks, in scale $k$, using every mask $M^{(k)}_{i,j}$ can get inpainted image $\hat{x}^{(k)}_{i,j}$, thus distance map $D(x, \hat{x}^{(k)}_{i,j})$ can be obtained according to \eqref{eq: distance}. For the scale $k$, combining masks for different patches, we can get a score map
\begin{equation}
    S^{(k)} = \sum_{i,j} M^{(k)}_{i,j} \odot D(x, \hat{x}^{(k)}_{i,j}).
\end{equation}
Aggregating different scale anomaly maps by harmonic mean, we can get the final anomaly map from multi-scale masks:
\begin{equation}
    S_{ms} = |\mathcal{K}| \left( \sum_{k \in \mathcal{K}} \frac{1}{S^{(k)}} \right)^{-1}.
\end{equation}

For the prototype-guided mask $M^{(p)}$ and inpainted image $\hat{x}^{(p)}$, we can also get anomaly score according to \eqref{eq: distance} and mask $M^{(p)}$:
\begin{equation}
    S_{pg} = M^{(p)} \odot D(x, \hat{x}^{(p)}).
\end{equation}

Finally, we can get the final anomaly score map by the combination of $S_{ms}$ and $S_{pg}$:
\begin{equation}
    S_{map} = (1-\alpha)S_{ms} + \alpha S_{pg}.
\end{equation}
For the image-level anomaly score, we use the maximum scores in $S_{map}$, thus we get $S_I$.

\section{Experiment}

\begin{table*}[ht]
\caption{Anomaly Detection and Localization results under k-shot setting on MVTec-AD and VisA benchmarks. }
\label{tb:perforamnce}
\centering
\addtolength{\tabcolsep}{-2pt}
\resizebox{0.9\textwidth}{!}{
\begin{tabular}{clcccccccccc}
\toprule
\multirow{3}{*}{Setup} & \multirow{3}{*}{Method} & \multicolumn{5}{c}{MVTec-AD}                     & \multicolumn{5}{c}{VisA} \\ 
\cmidrule(l{1em}r{1em}){3-7} \cmidrule(l{1em}r{1em}){8-12}
& & \multicolumn{3}{c}{image level} & \multicolumn{2}{c}{pixel level} & \multicolumn{3}{c}{image level} & \multicolumn{2}{c}{pixel level} \\
\cmidrule(l{1em}r{1em}){3-5} \cmidrule(l{1em}r{1em}){6-7} \cmidrule(l{1em}r{1em}){8-10} \cmidrule(l{1em}r{1em}){11-12}
&  & AUROC          & AUPR           & F1-max         & AUROC  & PRO   & AUROC          & AUPR         & F1-max         & AUROC & PRO \\ 
\midrule
\multirow{2}{*}{0-shot} & SD~\cite{ldm}   & 52.3 & 77.9 & 84.4 & 74.2 & 55.3  & 55.7 &  62.5 &  72.7 & 75.4 & 49.2 \\ 
    & WinCLIP~\cite{winclip}                    & \textbf{91.8} & \textbf{96.5} & \textbf{92.9} & 85.1 & 64.6  & 78.1 & 81.2 & 79.0 & 79.6 & 56.8  \\
    \cmidrule{2-12} 
    & \textbf{AnomalySD (ours)} & 86.6 & 93.7 & 90.5 & \textbf{87.6} & \textbf{81.5} & \textbf{80.9} & \textbf{83.4} &\textbf{79.8}& \textbf{92.8} & \textbf{90.5} \\ \midrule
\multirow{4}{*}{1-shot} & SPADE~\cite{spade} & 81.0 & 90.6 & 90.3 & 91.2 & 83.9  & 79.5 & 82.0 & 80.7 & 95.6 & 84.1 \\
    & PaDiM~\cite{padim}                     & 76.6 & 88.1 & 88.2 & 89.3 & 73.3  & 62.8 & 68.3 & 75.3 & 89.9 & 64.3 \\
    & PatchCore~\cite{patchcore}             & 83.4 & 92.2 & 90.5 & 92.0 & 79.7 & 79.9 & 82.8 & 81.7 & 95.4 & 80.5 \\
    & WinCLIP~\cite{winclip} & 93.1 & 96.5 & 93.7 & \textbf{95.2} & 87.1  & 83.8 & 85.1 & 83.1 & 96.4 & 85.1 \\
    \cmidrule{2-12} 
    & \textbf{AnomalySD (ours)} &    \textbf{93.6}  & \textbf{96.9} & \textbf{94.9}  & 94.8 & \textbf{89.2} & \textbf{86.1}    & \textbf{89.1}      &  \textbf{83.4} & \textbf{96.5} & \textbf{93.9}     \\ 
    \midrule
\multirow{4}{*}{2-shot} & SPADE~\cite{spade} & 82.9 & 91.7 & 91.1 & 92.0 & 85.7 & 80.7 & 82.3 & 81.7 & 96.2 & 85.7  \\
                        & PaDiM~\cite{padim}& 78.9 & 89.3 & 89.2 & 91.3 & 78.2 & 67.4 & 71.6 & 75.7 & 92.0 & 70.1  \\
     & PatchCore~\cite{patchcore}    & 86.3 & 93.8 & 92.0 & 93.3 & 82.3  & 81.6   & 84.8  & 82.5 & 96.1 & 82.6    \\
     & WinCLIP~\cite{winclip} & 94.4 & 97.0 & 94.4 & \textbf{96.0} & 88.4 & 84.6 & 85.8 & 83.0 & 96.8 & 86.2 \\
      \cmidrule{2-12} 
     & \textbf{AnomalySD (ours)}  &    \textbf{94.8} & \textbf{97.0} & \textbf{95.2}  & 95.8 & \textbf{90.4} &  \textbf{87.4}  & \textbf{90.1}  &  \textbf{83.7} & \textbf{96.8} & \textbf{94.1}  \\ \midrule
\multirow{4}{*}{4-shot} & SPADE~\cite{spade}    & 84.8      & 92.5      & 91.5  & 92.7 & 87.0 & 81.7   & 83.4  & 82.1  & 96.6 & 87.3 \\
     & PaDiM~\cite{padim}    & 80.4      & 90.5      & 90.2  & 92.6 & 81.3  & 72.8   & 75.6  & 78.0 & 93.2 & 72.6   \\
     & PatchCore~\cite{patchcore}   & 88.8      & 94.5      & 92.6  & 94.3 & 84.3  & 85.3   & 87.5  & 84.3 & 96.8 & 84.9  \\
      & WinCLIP~\cite{winclip} & 95.2 & 97.3 & 94.7 &  \textbf{96.2} & 89.0 & 87.3 & 88.8 & 84.2 & 97.2 & 87.6 \\
     \cmidrule{2-12} 
     & \textbf{AnomalySD (ours)}   &    \textbf{95.6} & \textbf{97.6} & \textbf{95.6}  & \textbf{96.2} & \textbf{90.8} & \textbf{88.9}  &  \textbf{90.9}  &  \textbf{85.4} & \textbf{97.5} & \textbf{94.3} \\ 
\bottomrule
\end{tabular}}
\end{table*}

\subsection{Experimental Setups}
\paragraph{\textbf{Datasets}} Our experiments are conducted on the MVTec-AD and VisA datasets, which simulate real-world industrial anomaly detection scenarios. The MVTec-AD~\cite{mvtec} dataset consists of 10 object categories and 5 texture categories. The training set contains 3629 normal samples, while the test set comprises 1725 images with various anomaly types, including both normal and anomalous samples. 
The VisA dataset~\cite{visa} comprises 10,821 high-resolution images, including 9,621 normal images and 1,200 abnormal images with 78 different types of anomalies. The dataset includes 12 distinct categories, broadly categorized into complex textures, multiple objects, and single objects.

\paragraph{\textbf{Evaluation metrics}}
Referring to previous work, in image-level anomaly detection, we utilize metrics such as the Area Under the Receiver Operating Characteristic Curve (AUROC), Average Precision (AUPR), and F1-max for better evaluation under situation of data imbalance. For pixel-level anomaly localization, we employ pixel-wise AUROC and F1-max, along with Per-Region Overlap (PRO) scores.

\paragraph{\textbf{Implementation Details}}
All samples from MVTec-AD and VisA datasets are scaled to $512 \times 512$. 
In the fine-tuning stage, data augmentations such as adjusting contrast brightness, scaling, and rotation are applied to the training samples to increase the diversity of few-shot samples. For the threshold $\gamma$ used for IoU of training mask $m$ and object foreground $m_f$, we randomly set it to 0.0, 0.2, and 0.5 with equal probability.
Adam optimizer is used and the learning rate is set to $1e^{-4}$.
We train the denoising network for 4000 epochs and the decoder for 200 epochs on NVIDIA GeForce RTX 3090 with a batch size of 8. 
In the inference stage, the denoising step is set to 50, and the noise strength $\lambda$ is specified for each category because the details of them are different. For the prototype-guided masks, we employ EfficientNet B6 as the feature extractor, and during the test we set the weight $\alpha$ to 0.1 for multi-scale and prototype score map fusion. For the anomaly map, we smooth it by using a Gaussian blur filter with $\sigma=4$.

\subsection{Anomaly Detection and Localization Results}
\paragraph{\textbf{Few-shot anomaly classification and segmentation}} 
The results of anomaly detection and localization under zero-shot and few-normal-shot settings are in Table \ref{tb:perforamnce}.  In the table, we compared our experimental results with previous works on the average performance of all classes in MVTec and VisA. For zero-shot learning, the multi-scale masks and noise strength control proposed in our approach can also be employed, thus we compare with the original SD with designed text prompts to show the effectiveness of these proposed modules. It can be observed that even without fine-tuning, our method still significantly outperforms the original SD. In other few-shot settings, our results show the competitive overall performance on both industrial datasets. On the MVTec dataset, the localization of AnomalySD is slightly inferior to WinCLIP because it uses prompts to describe the anomalous state. However, on the VisA dataset, AnomalySD surpasses WinCLIP a lot, by 2.3\%, 2.8\%, and 1.6\% in AUC in image level for 1, 2, and 4 shots respectively. The reason can be the hard-designed prompts of anomalous state used in WinCLIP are not very suitable for objects in VisA, but AnomalySD does not need such prompts to describe anomalous state. Meanwhile, we got generally better performance on 4-shot setting on both MVTec-AD and VisA datasets.
Comparing the performance of different shots, AnomalySD exhibits better performance when adding the shots of normal samples, highlighting the positive impact of fine-tuning SD with few-shot normal images on enhancing inpainting capability for anomaly detection and localization.

\paragraph{\textbf{Many-shot anomaly classification and segmentation}} 
In Table \ref{table:ac-many}, we compare our few-shot results with some prior many-shot works on MVTec-AD. It can be observed that our 1-shot AnomalySD outperforms recent few-shot methods such as TDG and DiffNet, even though their results are achieved with more than 10 shots. In particular, our 1-shot results surpass the aggregated 4-shot performance of RegAD, which uses abundant images from other categories but strictly limits the $k$-shots settings on the target category for training.
Furthermore, although all of our experiments are conducted in a multi-class setting without specific training for each category, our results still significantly outperform the few-shot works of bespoke training for each category individually. 

\begin{table}[h]
\caption{Comparison of many-shot methods in image and pixel AUROC on MVTec-AD}
% \vspace{-0.2cm}
\label{table:ac-many}
\begin{tabular}{lccc}
\toprule
Method        & Setup     & i-AUROC   & p-AUROC   \\ \midrule
\multirow{3}{*}{AnomalySD  (ours)} & 1-shot    & 93.6     &   94.8   \\
 & 2-shot    & 94.8     &   95.8   \\
 & 4-shot    & \textbf{95.6}     &   \textbf{96.2}   \\ \midrule
DiffNet~\cite{diffnet}       & 16-shot   & 87.3 & -    \\
TDG~\cite{tdg}           & 10-shot   & 78.0 & -    \\
RegAD~\cite{regad}         & 2-shot    & 81.5 & 93.3 \\
RegAD~\cite{regad}         & 4+agg.    & 88.1 & 95.8 \\
\bottomrule
\end{tabular}
% \vspace{-0.2cm}
\end{table}

\begin{figure}[ht]
% \vskip 0.2in
% \begin{center}
\centering
\includegraphics[width=\columnwidth]{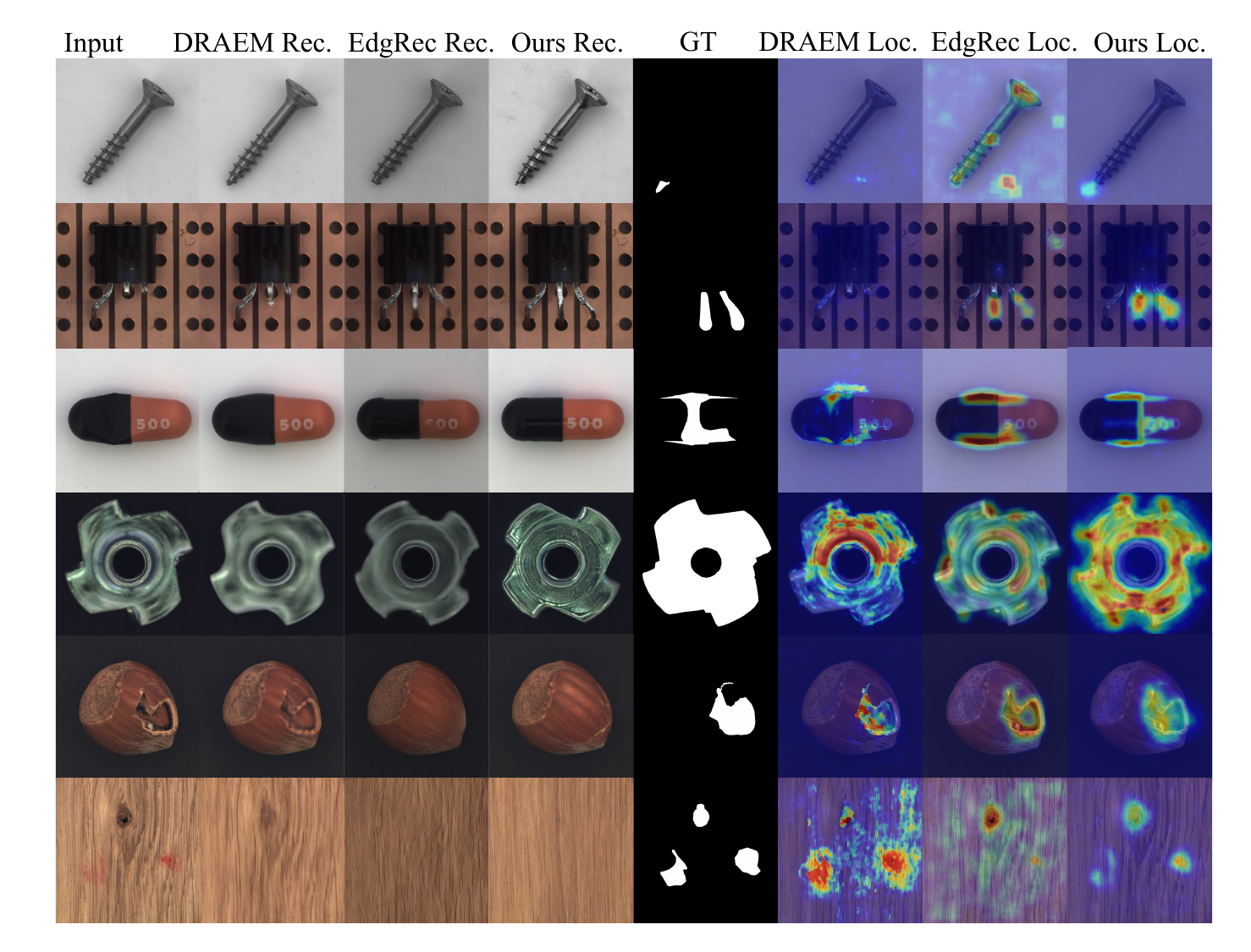}
% \vspace{-0.2cm}
\caption{
Comparison of the reconstruction results and location results between our 1-shot method with full-shot methods DRAEM~\cite{draem} and EdgRec~\cite{EdgRec}.}
\label{fig.rec_loc}
% \end{center}
% \vskip -0.2in
% \vspace{-0.2cm}
\end{figure}

\paragraph{\textbf{Visualization results}} We conducted extensive qualitative studies on the MVTec-AD and VisA datasets, demonstrating the superiority of our method in terms of image reconstruction and anomaly localization. As shown in Fig.\ref{fig.rec_loc}, we compared the reconstruction and anomaly localization results of abnormal samples with the full normal supervised method DRAEM~\cite{draem} and EdgRec~\cite{EdgRec}. Our approach exhibits superior reconstruction capabilities for abnormal regions, enabling more accurate localization compared to current reconstructive methods.

\subsection{Ablation Study}
All of the ablation experiments are conducted on the MVTec dataset under 1-shot setting, we report the performance of image-level AUROC(i-AUROC) and pixel-level AUROC(p-AUROC).

\begin{table}[h]
% \vspace{-0.1cm}
\caption{Ablation of fine-tuning modules of decoder $\mathcal{D}(\cdot)$ and denoising network $\epsilon_{\theta}(\cdot)$.}
\vspace{-0.1cm}
\label{tab:ablation_fint_tune}
\begin{tabular}{cccc}
\toprule
\multicolumn{2}{c}{Fine-tuned module} & \multicolumn{2}{c}{Performance} \\  
\cmidrule(lr){1-2} \cmidrule(lr){3-4}
$\mathcal{D}(\cdot)$& $\epsilon_{\theta}(\cdot)$ & i-AUROC & p-AUROC \\
\midrule
            &               &  86.6       &  87.6        \\
\checkmark  &               &  87.2       &  88.9        \\
            &  \checkmark   &  93.2       &  94.1        \\
\checkmark  &  \checkmark   &  \textbf{93.6}       &  \textbf{94.8}        \\
\bottomrule
\end{tabular}
% \vspace{-0.1cm}
\end{table}

\begin{table}[h]
\caption{Ablation of mask design and prompt design.}
% \vspace{-0.1cm}
\label{tab:design}
\centering
\addtolength{\tabcolsep}{-2pt}
\resizebox{0.48\textwidth}{!}{
\begin{tabular}{cccccc}
\toprule
\multirow{2}{*}{Performance} & \multicolumn{2}{c}{Mask} & \multicolumn{3}{c}{Prompt} \\
\cmidrule(lr){2-3} \cmidrule(lr){4-6} 
& random & mask-design & none & simple & prompt-design \\
\midrule
i-AUROC & 93.0  &  \textbf{93.6} & 91.5 & 92.4  & \textbf{93.6} \\
p-AUROC & 93.9  &  \textbf{94.8} & 92.7 & 93.3  & \textbf{94.8} \\ 
\bottomrule
\end{tabular}}
\end{table}

\begin{table}[h]
\caption{Ablation studies of masks in inference stage. }
% \vspace{-0.1cm}
\label{tab:mask_infer}
\begin{tabular}{lcc}
\toprule
Mask & i-AUROC & p-AUROC \\  
\midrule
one full mask               &  84.9     &   89.6    \\
multi-scale                 &  93.1     &   94.2    \\ 
multi-scale + prototype     &  \textbf{93.6}     &   \textbf{94.8}    \\
\bottomrule
\end{tabular}
% \vspace{-0.2cm}
\end{table}

\begin{figure}[ht]
% \vskip 0.2in
% \begin{center}
\centering
\includegraphics[width=0.9\columnwidth]{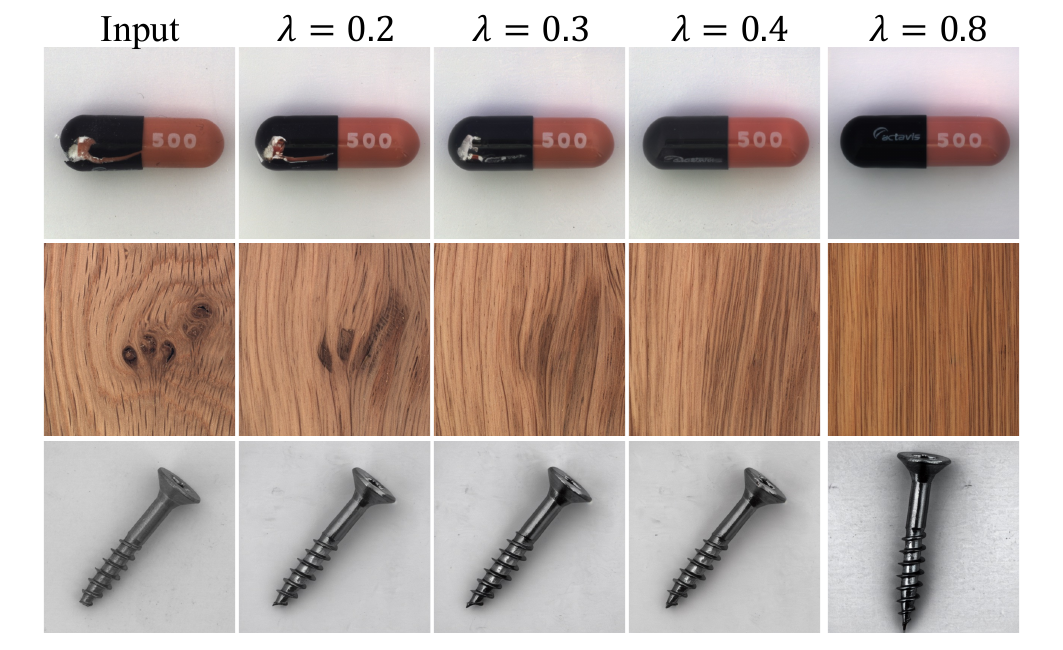}
\caption{
Ablations of noise strength $\lambda$ in inference stage.}
% \vspace{-0.2cm}
\label{fig.guidance}
% \end{center}
% \vskip -0.2in
\end{figure}

\paragraph{\textbf{Fine-tuned modules}} Table \ref{tab:ablation_fint_tune} illustrates the performance of fine-tuning different modules. Compared to no fine-tuned SD, fine-tuning the decoder can improve the AUROC of image and pixel by 0.6\% and 1.3\%, because the fine-tuned decoder can better recover pixels from latent space in detail.
% , and measure the distance between input image and reconstructed image more accurately, promoting anomaly localization.
Fine-tuning the denoising network can greatly improve the AUROC of image and pixel by 5.6\% and 6.5\% because it is trained for learning to re-establish normal patterns on the provided few-shot dataset. When both modules are fine-tuned, the AUROC of image and pixel can be improved by 6\% and 7.2\%.

\paragraph{\textbf{Mask and prompt designs for fine-tuning}} Table \ref{tab:design} shows the performance of different mask and prompt designs for fine-tuning SD. For the mask, we compare the proposed mask design with randomly generated mask, finding the proposed mask design can surpass it by 0.6\%, and 0.9\% for i-AUROC, and p-AUROC respectively. For the prompt, compared to the two strategies: no prompt, a simple design in the form of ``\texttt{A photo of [c].}'', our proposed prompt design surpasses the simple prompt by 1.2\%, 1.5\% for i-AUROC and p-AUROC respectively. The good detection and localization performance proves the effectiveness of prompt design.

\paragraph{\textbf{Masks in the inference stage}} We ablate on different masks used for inpainting anomalous areas in the inference stage. In Table \ref{tab:mask_infer}, one full mask covers the whole image, compared to which, multi-scale can improve i-AUROC and p-AUROC by 8.2\% and 4.6\% respectively. Adding prototype-guided masks, the detection and localization performance can be further improved, so we adopt multi-scale and prototype-guided masks in the inference stage.

\paragraph{\textbf{Noise strength for inpainting}} As mentioned in Sec \ref{sec:infer}, if we sample noise from $\mathcal{N}(0,I)$ at the beginning of the inference stage, the inpainted image can exhibit different patterns from the original image. As shown in Figure \ref{fig.guidance}, under the noise strength $\lambda=0.8$, the orientation of inpainted screw is different from the input screw, leading to deviations in localizing anomalies. So the controlled noise strength can not be too large. Nor the strength can be too small like $\lambda=0.2$ shown in Figure \ref{fig.guidance}, under which we can not inapint anomalous area into normal. $\lambda = 0.4$ is an appropriate choice for capsule, wood, and screw.

\subsection{Conclusion}
We propose a SD-based framework AnomalySD, which detects and localizes anomalies under few shot and multi-class settings. We introduce a combination of hierarchical text prompt and mask designs to adapt SD for anomaly detection and use multi-scale masks, prototype-guided masks to mask anomalies and restore them into normal patterns via finetuned SD. Our approach achieves competitive performance on the MVTec-AD and VisA datasets for few-shot multi-class anomaly detection. 
For further improvement, adaptive prompt learning and noise guidance is a promising direction to reduce the reliance on manually set prior information and transition to a model-adaptive learning process.

%%
%% The next two lines define the bibliography style to be used, and
%% the bibliography file.
\bibliographystyle{ACM-Reference-Format}
\bibliography{main}

%%% -*-BibTeX-*-
%%% Do NOT edit. File created by BibTeX with style
%%% ACM-Reference-Format-Journals [18-Jan-2012].

\begin{thebibliography}{42}

%%% ====================================================================
%%% NOTE TO THE USER: you can override these defaults by providing
%%% customized versions of any of these macros before the \bibliography
%%% command.  Each of them MUST provide its own final punctuation,
%%% except for \shownote{}, \showDOI{}, and \showURL{}.  The latter two
%%% do not use final punctuation, in order to avoid confusing it with
%%% the Web address.
%%%
%%% To suppress output of a particular field, define its macro to expand
%%% to an empty string, or better, \unskip, like this:
%%%
%%% \newcommand{\showDOI}[1]{\unskip}   % LaTeX syntax
%%%
%%% \def \showDOI #1{\unskip}           % plain TeX syntax
%%%
%%% ====================================================================

\ifx \showCODEN    \undefined \def \showCODEN     #1{\unskip}     \fi
\ifx \showDOI      \undefined \def \showDOI       #1{#1}\fi
\ifx \showISBNx    \undefined \def \showISBNx     #1{\unskip}     \fi
\ifx \showISBNxiii \undefined \def \showISBNxiii  #1{\unskip}     \fi
\ifx \showISSN     \undefined \def \showISSN      #1{\unskip}     \fi
\ifx \showLCCN     \undefined \def \showLCCN      #1{\unskip}     \fi
\ifx \shownote     \undefined \def \shownote      #1{#1}          \fi
\ifx \showarticletitle \undefined \def \showarticletitle #1{#1}   \fi
\ifx \showURL      \undefined \def \showURL       {\relax}        \fi
% The following commands are used for tagged output and should be
% invisible to TeX
\providecommand\bibfield[2]{#2}
\providecommand\bibinfo[2]{#2}
\providecommand\natexlab[1]{#1}
\providecommand\showeprint[2][]{arXiv:#2}

\bibitem[Akcay et~al\mbox{.}(2019)]%
        {ganomaly}
\bibfield{author}{\bibinfo{person}{Samet Akcay}, \bibinfo{person}{Amir Atapour-Abarghouei}, {and} \bibinfo{person}{Toby~P Breckon}.} \bibinfo{year}{2019}\natexlab{}.
\newblock \showarticletitle{Ganomaly: Semi-supervised anomaly detection via adversarial training}. In \bibinfo{booktitle}{\emph{Computer Vision--ACCV 2018: 14th Asian Conference on Computer Vision, Perth, Australia, December 2--6, 2018, Revised Selected Papers, Part III 14}}. Springer, \bibinfo{pages}{622--637}.
\newblock


\bibitem[Ak{\c{c}}ay et~al\mbox{.}(2019)]%
        {skipgan}
\bibfield{author}{\bibinfo{person}{Samet Ak{\c{c}}ay}, \bibinfo{person}{Amir Atapour-Abarghouei}, {and} \bibinfo{person}{Toby~P Breckon}.} \bibinfo{year}{2019}\natexlab{}.
\newblock \showarticletitle{Skip-ganomaly: Skip connected and adversarially trained encoder-decoder anomaly detection}. In \bibinfo{booktitle}{\emph{2019 International Joint Conference on Neural Networks (IJCNN)}}. IEEE, \bibinfo{pages}{1--8}.
\newblock


\bibitem[Bergmann et~al\mbox{.}(2019)]%
        {mvtec}
\bibfield{author}{\bibinfo{person}{Paul Bergmann}, \bibinfo{person}{Michael Fauser}, \bibinfo{person}{David Sattlegger}, {and} \bibinfo{person}{Carsten Steger}.} \bibinfo{year}{2019}\natexlab{}.
\newblock \showarticletitle{MVTec AD--A comprehensive real-world dataset for unsupervised anomaly detection}. In \bibinfo{booktitle}{\emph{Proceedings of the IEEE/CVF conference on computer vision and pattern recognition}}. \bibinfo{pages}{9592--9600}.
\newblock


\bibitem[Bergmann et~al\mbox{.}(2020)]%
        {uninformed}
\bibfield{author}{\bibinfo{person}{Paul Bergmann}, \bibinfo{person}{Michael Fauser}, \bibinfo{person}{David Sattlegger}, {and} \bibinfo{person}{Carsten Steger}.} \bibinfo{year}{2020}\natexlab{}.
\newblock \showarticletitle{Uninformed students: Student-teacher anomaly detection with discriminative latent embeddings}. In \bibinfo{booktitle}{\emph{Proceedings of the IEEE/CVF conference on computer vision and pattern recognition}}. \bibinfo{pages}{4183--4192}.
\newblock


\bibitem[Cohen and Hoshen(2020)]%
        {spade}
\bibfield{author}{\bibinfo{person}{Niv Cohen} {and} \bibinfo{person}{Yedid Hoshen}.} \bibinfo{year}{2020}\natexlab{}.
\newblock \showarticletitle{Sub-image anomaly detection with deep pyramid correspondences}.
\newblock \bibinfo{journal}{\emph{arXiv preprint arXiv:2005.02357}} (\bibinfo{year}{2020}).
\newblock


\bibitem[Defard et~al\mbox{.}(2021)]%
        {padim}
\bibfield{author}{\bibinfo{person}{Thomas Defard}, \bibinfo{person}{Aleksandr Setkov}, \bibinfo{person}{Angelique Loesch}, {and} \bibinfo{person}{Romaric Audigier}.} \bibinfo{year}{2021}\natexlab{}.
\newblock \showarticletitle{Padim: a patch distribution modeling framework for anomaly detection and localization}. In \bibinfo{booktitle}{\emph{International Conference on Pattern Recognition}}. Springer, \bibinfo{pages}{475--489}.
\newblock


\bibitem[Dehaene et~al\mbox{.}(2020)]%
        {iterative}
\bibfield{author}{\bibinfo{person}{David Dehaene}, \bibinfo{person}{Oriel Frigo}, \bibinfo{person}{S{\'e}bastien Combrexelle}, {and} \bibinfo{person}{Pierre Eline}.} \bibinfo{year}{2020}\natexlab{}.
\newblock \showarticletitle{Iterative energy-based projection on a normal data manifold for anomaly localization}.
\newblock \bibinfo{journal}{\emph{arXiv preprint arXiv:2002.03734}} (\bibinfo{year}{2020}).
\newblock


\bibitem[Deng and Li(2022)]%
        {distillation}
\bibfield{author}{\bibinfo{person}{Hanqiu Deng} {and} \bibinfo{person}{Xingyu Li}.} \bibinfo{year}{2022}\natexlab{}.
\newblock \showarticletitle{Anomaly detection via reverse distillation from one-class embedding}. In \bibinfo{booktitle}{\emph{Proceedings of the IEEE/CVF Conference on Computer Vision and Pattern Recognition}}. \bibinfo{pages}{9737--9746}.
\newblock


\bibitem[Deng et~al\mbox{.}(2009)]%
        {imagenet}
\bibfield{author}{\bibinfo{person}{Jia Deng}, \bibinfo{person}{Wei Dong}, \bibinfo{person}{Richard Socher}, \bibinfo{person}{Li-Jia Li}, \bibinfo{person}{Kai Li}, {and} \bibinfo{person}{Li Fei-Fei}.} \bibinfo{year}{2009}\natexlab{}.
\newblock \showarticletitle{Imagenet: A large-scale hierarchical image database}. In \bibinfo{booktitle}{\emph{2009 IEEE conference on computer vision and pattern recognition}}. Ieee, \bibinfo{pages}{248--255}.
\newblock


\bibitem[Gong et~al\mbox{.}(2019)]%
        {memae}
\bibfield{author}{\bibinfo{person}{Dong Gong}, \bibinfo{person}{Lingqiao Liu}, \bibinfo{person}{Vuong Le}, \bibinfo{person}{Budhaditya Saha}, \bibinfo{person}{Moussa~Reda Mansour}, \bibinfo{person}{Svetha Venkatesh}, {and} \bibinfo{person}{Anton van~den Hengel}.} \bibinfo{year}{2019}\natexlab{}.
\newblock \showarticletitle{Memorizing normality to detect anomaly: Memory-augmented deep autoencoder for unsupervised anomaly detection}. In \bibinfo{booktitle}{\emph{Proceedings of the IEEE/CVF International Conference on Computer Vision}}. \bibinfo{pages}{1705--1714}.
\newblock


\bibitem[Goodfellow et~al\mbox{.}(2014)]%
        {gan}
\bibfield{author}{\bibinfo{person}{Ian Goodfellow}, \bibinfo{person}{Jean Pouget-Abadie}, \bibinfo{person}{Mehdi Mirza}, \bibinfo{person}{Bing Xu}, \bibinfo{person}{David Warde-Farley}, \bibinfo{person}{Sherjil Ozair}, \bibinfo{person}{Aaron Courville}, {and} \bibinfo{person}{Yoshua Bengio}.} \bibinfo{year}{2014}\natexlab{}.
\newblock \showarticletitle{Generative adversarial nets}.
\newblock \bibinfo{journal}{\emph{Advances in neural information processing systems}}  \bibinfo{volume}{27} (\bibinfo{year}{2014}).
\newblock


\bibitem[He et~al\mbox{.}(2024)]%
        {diad}
\bibfield{author}{\bibinfo{person}{Haoyang He}, \bibinfo{person}{Jiangning Zhang}, \bibinfo{person}{Hongxu Chen}, \bibinfo{person}{Xuhai Chen}, \bibinfo{person}{Zhishan Li}, \bibinfo{person}{Xu Chen}, \bibinfo{person}{Yabiao Wang}, \bibinfo{person}{Chengjie Wang}, {and} \bibinfo{person}{Lei Xie}.} \bibinfo{year}{2024}\natexlab{}.
\newblock \showarticletitle{A Diffusion-Based Framework for Multi-Class Anomaly Detection}. In \bibinfo{booktitle}{\emph{Proceedings of the AAAI Conference on Artificial Intelligence}}, Vol.~\bibinfo{volume}{38}. \bibinfo{pages}{8472--8480}.
\newblock


\bibitem[Ho et~al\mbox{.}(2020)]%
        {ddpm}
\bibfield{author}{\bibinfo{person}{Jonathan Ho}, \bibinfo{person}{Ajay Jain}, {and} \bibinfo{person}{Pieter Abbeel}.} \bibinfo{year}{2020}\natexlab{}.
\newblock \showarticletitle{Denoising diffusion probabilistic models}.
\newblock \bibinfo{journal}{\emph{Advances in neural information processing systems}}  \bibinfo{volume}{33} (\bibinfo{year}{2020}), \bibinfo{pages}{6840--6851}.
\newblock


\bibitem[Hou et~al\mbox{.}(2021)]%
        {dividegan}
\bibfield{author}{\bibinfo{person}{Jinlei Hou}, \bibinfo{person}{Yingying Zhang}, \bibinfo{person}{Qiaoyong Zhong}, \bibinfo{person}{Di Xie}, \bibinfo{person}{Shiliang Pu}, {and} \bibinfo{person}{Hong Zhou}.} \bibinfo{year}{2021}\natexlab{}.
\newblock \showarticletitle{Divide-and-assemble: Learning block-wise memory for unsupervised anomaly detection}. In \bibinfo{booktitle}{\emph{Proceedings of the IEEE/CVF International Conference on Computer Vision}}. \bibinfo{pages}{8791--8800}.
\newblock


\bibitem[Hu et~al\mbox{.}(2023)]%
        {diffusiongeneration}
\bibfield{author}{\bibinfo{person}{Teng Hu}, \bibinfo{person}{Jiangning Zhang}, \bibinfo{person}{Ran Yi}, \bibinfo{person}{Yuzhen Du}, \bibinfo{person}{Xu Chen}, \bibinfo{person}{Liang Liu}, \bibinfo{person}{Yabiao Wang}, {and} \bibinfo{person}{Chengjie Wang}.} \bibinfo{year}{2023}\natexlab{}.
\newblock \showarticletitle{AnomalyDiffusion: Few-Shot Anomaly Image Generation with Diffusion Model}.
\newblock \bibinfo{journal}{\emph{arXiv preprint arXiv:2312.05767}} (\bibinfo{year}{2023}).
\newblock


\bibitem[Huang et~al\mbox{.}(2022)]%
        {regad}
\bibfield{author}{\bibinfo{person}{Chaoqin Huang}, \bibinfo{person}{Haoyan Guan}, \bibinfo{person}{Aofan Jiang}, \bibinfo{person}{Ya Zhang}, \bibinfo{person}{Michael Spratling}, {and} \bibinfo{person}{Yan-Feng Wang}.} \bibinfo{year}{2022}\natexlab{}.
\newblock \showarticletitle{Registration based few-shot anomaly detection}. In \bibinfo{booktitle}{\emph{European Conference on Computer Vision}}. Springer, \bibinfo{pages}{303--319}.
\newblock


\bibitem[Jeong et~al\mbox{.}(2023)]%
        {winclip}
\bibfield{author}{\bibinfo{person}{Jongheon Jeong}, \bibinfo{person}{Yang Zou}, \bibinfo{person}{Taewan Kim}, \bibinfo{person}{Dongqing Zhang}, \bibinfo{person}{Avinash Ravichandran}, {and} \bibinfo{person}{Onkar Dabeer}.} \bibinfo{year}{2023}\natexlab{}.
\newblock \showarticletitle{Winclip: Zero-/few-shot anomaly classification and segmentation}. In \bibinfo{booktitle}{\emph{Proceedings of the IEEE/CVF Conference on Computer Vision and Pattern Recognition}}. \bibinfo{pages}{19606--19616}.
\newblock


\bibitem[Li et~al\mbox{.}(2020)]%
        {smai}
\bibfield{author}{\bibinfo{person}{Zhenyu Li}, \bibinfo{person}{Ning Li}, \bibinfo{person}{Kaitao Jiang}, \bibinfo{person}{Zhiheng Ma}, \bibinfo{person}{Xing Wei}, \bibinfo{person}{Xiaopeng Hong}, {and} \bibinfo{person}{Yihong Gong}.} \bibinfo{year}{2020}\natexlab{}.
\newblock \showarticletitle{Superpixel Masking and Inpainting for Self-Supervised Anomaly Detection.}. In \bibinfo{booktitle}{\emph{Bmvc}}.
\newblock


\bibitem[Liang et~al\mbox{.}(2023)]%
        {ocrgan}
\bibfield{author}{\bibinfo{person}{Yufei Liang}, \bibinfo{person}{Jiangning Zhang}, \bibinfo{person}{Shiwei Zhao}, \bibinfo{person}{Runze Wu}, \bibinfo{person}{Yong Liu}, {and} \bibinfo{person}{Shuwen Pan}.} \bibinfo{year}{2023}\natexlab{}.
\newblock \showarticletitle{Omni-frequency channel-selection representations for unsupervised anomaly detection}.
\newblock \bibinfo{journal}{\emph{IEEE Transactions on Image Processing}} (\bibinfo{year}{2023}).
\newblock


\bibitem[Liu et~al\mbox{.}(2022)]%
        {EdgRec}
\bibfield{author}{\bibinfo{person}{Tongkun Liu}, \bibinfo{person}{Bing Li}, \bibinfo{person}{Zhuo Zhao}, \bibinfo{person}{Xiao Du}, \bibinfo{person}{Bingke Jiang}, {and} \bibinfo{person}{Leqi Geng}.} \bibinfo{year}{2022}\natexlab{}.
\newblock \showarticletitle{Reconstruction from edge image combined with color and gradient difference for industrial surface anomaly detection}.
\newblock \bibinfo{journal}{\emph{arXiv preprint arXiv:2210.14485}} (\bibinfo{year}{2022}).
\newblock


\bibitem[Liu et~al\mbox{.}(2020)]%
        {towards}
\bibfield{author}{\bibinfo{person}{Wenqian Liu}, \bibinfo{person}{Runze Li}, \bibinfo{person}{Meng Zheng}, \bibinfo{person}{Srikrishna Karanam}, \bibinfo{person}{Ziyan Wu}, \bibinfo{person}{Bir Bhanu}, \bibinfo{person}{Richard~J Radke}, {and} \bibinfo{person}{Octavia Camps}.} \bibinfo{year}{2020}\natexlab{}.
\newblock \showarticletitle{Towards visually explaining variational autoencoders}. In \bibinfo{booktitle}{\emph{Proceedings of the IEEE/CVF Conference on Computer Vision and Pattern Recognition}}. \bibinfo{pages}{8642--8651}.
\newblock


\bibitem[Radford et~al\mbox{.}(2021)]%
        {clip}
\bibfield{author}{\bibinfo{person}{Alec Radford}, \bibinfo{person}{Jong~Wook Kim}, \bibinfo{person}{Chris Hallacy}, \bibinfo{person}{Aditya Ramesh}, \bibinfo{person}{Gabriel Goh}, \bibinfo{person}{Sandhini Agarwal}, \bibinfo{person}{Girish Sastry}, \bibinfo{person}{Amanda Askell}, \bibinfo{person}{Pamela Mishkin}, \bibinfo{person}{Jack Clark}, {et~al\mbox{.}}} \bibinfo{year}{2021}\natexlab{}.
\newblock \showarticletitle{Learning transferable visual models from natural language supervision}. In \bibinfo{booktitle}{\emph{International conference on machine learning}}. PMLR, \bibinfo{pages}{8748--8763}.
\newblock


\bibitem[Rombach et~al\mbox{.}(2022)]%
        {ldm}
\bibfield{author}{\bibinfo{person}{Robin Rombach}, \bibinfo{person}{Andreas Blattmann}, \bibinfo{person}{Dominik Lorenz}, \bibinfo{person}{Patrick Esser}, {and} \bibinfo{person}{Bj{\"o}rn Ommer}.} \bibinfo{year}{2022}\natexlab{}.
\newblock \showarticletitle{High-resolution image synthesis with latent diffusion models}. In \bibinfo{booktitle}{\emph{Proceedings of the IEEE/CVF conference on computer vision and pattern recognition}}. \bibinfo{pages}{10684--10695}.
\newblock


\bibitem[Ronneberger et~al\mbox{.}(2015)]%
        {unet}
\bibfield{author}{\bibinfo{person}{Olaf Ronneberger}, \bibinfo{person}{Philipp Fischer}, {and} \bibinfo{person}{Thomas Brox}.} \bibinfo{year}{2015}\natexlab{}.
\newblock \showarticletitle{U-net: Convolutional networks for biomedical image segmentation}. In \bibinfo{booktitle}{\emph{Medical Image Computing and Computer-Assisted Intervention--MICCAI 2015: 18th International Conference, Munich, Germany, October 5-9, 2015, Proceedings, Part III 18}}. Springer, \bibinfo{pages}{234--241}.
\newblock


\bibitem[Roth et~al\mbox{.}(2022)]%
        {patchcore}
\bibfield{author}{\bibinfo{person}{Karsten Roth}, \bibinfo{person}{Latha Pemula}, \bibinfo{person}{Joaquin Zepeda}, \bibinfo{person}{Bernhard Sch{\"o}lkopf}, \bibinfo{person}{Thomas Brox}, {and} \bibinfo{person}{Peter Gehler}.} \bibinfo{year}{2022}\natexlab{}.
\newblock \showarticletitle{Towards total recall in industrial anomaly detection}. In \bibinfo{booktitle}{\emph{Proceedings of the IEEE/CVF Conference on Computer Vision and Pattern Recognition}}. \bibinfo{pages}{14318--14328}.
\newblock


\bibitem[Rudolph et~al\mbox{.}(2021)]%
        {diffnet}
\bibfield{author}{\bibinfo{person}{Marco Rudolph}, \bibinfo{person}{Bastian Wandt}, {and} \bibinfo{person}{Bodo Rosenhahn}.} \bibinfo{year}{2021}\natexlab{}.
\newblock \showarticletitle{Same same but differnet: Semi-supervised defect detection with normalizing flows}. In \bibinfo{booktitle}{\emph{Proceedings of the IEEE/CVF winter conference on applications of computer vision}}. \bibinfo{pages}{1907--1916}.
\newblock


\bibitem[Rumelhart et~al\mbox{.}(1985)]%
        {ae}
\bibfield{author}{\bibinfo{person}{David~E Rumelhart}, \bibinfo{person}{Geoffrey~E Hinton}, \bibinfo{person}{Ronald~J Williams}, {et~al\mbox{.}}} \bibinfo{year}{1985}\natexlab{}.
\newblock \bibinfo{title}{Learning internal representations by error propagation}.
\newblock
\newblock


\bibitem[Schlegl et~al\mbox{.}(2019)]%
        {f-anogan}
\bibfield{author}{\bibinfo{person}{Thomas Schlegl}, \bibinfo{person}{Philipp Seeb{\"o}ck}, \bibinfo{person}{Sebastian~M Waldstein}, \bibinfo{person}{Georg Langs}, {and} \bibinfo{person}{Ursula Schmidt-Erfurth}.} \bibinfo{year}{2019}\natexlab{}.
\newblock \showarticletitle{f-AnoGAN: Fast unsupervised anomaly detection with generative adversarial networks}.
\newblock \bibinfo{journal}{\emph{Medical image analysis}}  \bibinfo{volume}{54} (\bibinfo{year}{2019}), \bibinfo{pages}{30--44}.
\newblock


\bibitem[Schlegl et~al\mbox{.}(2017)]%
        {anogan}
\bibfield{author}{\bibinfo{person}{Thomas Schlegl}, \bibinfo{person}{Philipp Seeb{\"o}ck}, \bibinfo{person}{Sebastian~M Waldstein}, \bibinfo{person}{Ursula Schmidt-Erfurth}, {and} \bibinfo{person}{Georg Langs}.} \bibinfo{year}{2017}\natexlab{}.
\newblock \showarticletitle{Unsupervised anomaly detection with generative adversarial networks to guide marker discovery}. In \bibinfo{booktitle}{\emph{International conference on information processing in medical imaging}}. Springer, \bibinfo{pages}{146--157}.
\newblock


\bibitem[Sheynin et~al\mbox{.}(2021)]%
        {tdg}
\bibfield{author}{\bibinfo{person}{Shelly Sheynin}, \bibinfo{person}{Sagie Benaim}, {and} \bibinfo{person}{Lior Wolf}.} \bibinfo{year}{2021}\natexlab{}.
\newblock \showarticletitle{A hierarchical transformation-discriminating generative model for few shot anomaly detection}. In \bibinfo{booktitle}{\emph{Proceedings of the IEEE/CVF International Conference on Computer Vision}}. \bibinfo{pages}{8495--8504}.
\newblock


\bibitem[Song et~al\mbox{.}(2020)]%
        {ddim}
\bibfield{author}{\bibinfo{person}{Jiaming Song}, \bibinfo{person}{Chenlin Meng}, {and} \bibinfo{person}{Stefano Ermon}.} \bibinfo{year}{2020}\natexlab{}.
\newblock \showarticletitle{Denoising diffusion implicit models}.
\newblock \bibinfo{journal}{\emph{arXiv preprint arXiv:2010.02502}} (\bibinfo{year}{2020}).
\newblock


\bibitem[Wyatt et~al\mbox{.}(2022)]%
        {anoddpm}
\bibfield{author}{\bibinfo{person}{Julian Wyatt}, \bibinfo{person}{Adam Leach}, \bibinfo{person}{Sebastian~M Schmon}, {and} \bibinfo{person}{Chris~G Willcocks}.} \bibinfo{year}{2022}\natexlab{}.
\newblock \showarticletitle{Anoddpm: Anomaly detection with denoising diffusion probabilistic models using simplex noise}. In \bibinfo{booktitle}{\emph{Proceedings of the IEEE/CVF Conference on Computer Vision and Pattern Recognition}}. \bibinfo{pages}{650--656}.
\newblock


\bibitem[Xie et~al\mbox{.}(2023)]%
        {graphcore}
\bibfield{author}{\bibinfo{person}{Guoyang Xie}, \bibinfo{person}{Jinbao Wang}, \bibinfo{person}{Jiaqi Liu}, \bibinfo{person}{Feng Zheng}, {and} \bibinfo{person}{Yaochu Jin}.} \bibinfo{year}{2023}\natexlab{}.
\newblock \showarticletitle{Pushing the limits of fewshot anomaly detection in industry vision: Graphcore}.
\newblock \bibinfo{journal}{\emph{arXiv preprint arXiv:2301.12082}} (\bibinfo{year}{2023}).
\newblock


\bibitem[Yan et~al\mbox{.}(2021)]%
        {scadn}
\bibfield{author}{\bibinfo{person}{Xudong Yan}, \bibinfo{person}{Huaidong Zhang}, \bibinfo{person}{Xuemiao Xu}, \bibinfo{person}{Xiaowei Hu}, {and} \bibinfo{person}{Pheng-Ann Heng}.} \bibinfo{year}{2021}\natexlab{}.
\newblock \showarticletitle{Learning semantic context from normal samples for unsupervised anomaly detection}. In \bibinfo{booktitle}{\emph{Proceedings of the AAAI conference on artificial intelligence}}, Vol.~\bibinfo{volume}{35}. \bibinfo{pages}{3110--3118}.
\newblock


\bibitem[Yao et~al\mbox{.}(2023)]%
        {oneforall}
\bibfield{author}{\bibinfo{person}{Xincheng Yao}, \bibinfo{person}{Chongyang Zhang}, \bibinfo{person}{Ruoqi Li}, \bibinfo{person}{Jun Sun}, {and} \bibinfo{person}{Zhenyu Liu}.} \bibinfo{year}{2023}\natexlab{}.
\newblock \showarticletitle{One-for-all: Proposal masked cross-class anomaly detection}. In \bibinfo{booktitle}{\emph{Proceedings of the AAAI Conference on Artificial Intelligence}}, Vol.~\bibinfo{volume}{37}. \bibinfo{pages}{4792--4800}.
\newblock


\bibitem[You et~al\mbox{.}(2022)]%
        {unified}
\bibfield{author}{\bibinfo{person}{Zhiyuan You}, \bibinfo{person}{Lei Cui}, \bibinfo{person}{Yujun Shen}, \bibinfo{person}{Kai Yang}, \bibinfo{person}{Xin Lu}, \bibinfo{person}{Yu Zheng}, {and} \bibinfo{person}{Xinyi Le}.} \bibinfo{year}{2022}\natexlab{}.
\newblock \showarticletitle{A unified model for multi-class anomaly detection}.
\newblock \bibinfo{journal}{\emph{Advances in Neural Information Processing Systems}}  \bibinfo{volume}{35} (\bibinfo{year}{2022}), \bibinfo{pages}{4571--4584}.
\newblock


\bibitem[Zavrtanik et~al\mbox{.}(2021a)]%
        {draem}
\bibfield{author}{\bibinfo{person}{Vitjan Zavrtanik}, \bibinfo{person}{Matej Kristan}, {and} \bibinfo{person}{Danijel Sko{\v{c}}aj}.} \bibinfo{year}{2021}\natexlab{a}.
\newblock \showarticletitle{Draem-a discriminatively trained reconstruction embedding for surface anomaly detection}. In \bibinfo{booktitle}{\emph{Proceedings of the IEEE/CVF International Conference on Computer Vision}}. \bibinfo{pages}{8330--8339}.
\newblock


\bibitem[Zavrtanik et~al\mbox{.}(2021b)]%
        {riad}
\bibfield{author}{\bibinfo{person}{Vitjan Zavrtanik}, \bibinfo{person}{Matej Kristan}, {and} \bibinfo{person}{Danijel Sko{\v{c}}aj}.} \bibinfo{year}{2021}\natexlab{b}.
\newblock \showarticletitle{Reconstruction by inpainting for visual anomaly detection}.
\newblock \bibinfo{journal}{\emph{Pattern Recognition}}  \bibinfo{volume}{112} (\bibinfo{year}{2021}), \bibinfo{pages}{107706}.
\newblock


\bibitem[Zhang et~al\mbox{.}(2023)]%
        {diffusionad}
\bibfield{author}{\bibinfo{person}{Hui Zhang}, \bibinfo{person}{Zheng Wang}, \bibinfo{person}{Zuxuan Wu}, {and} \bibinfo{person}{Yu-Gang Jiang}.} \bibinfo{year}{2023}\natexlab{}.
\newblock \bibinfo{title}{DiffusionAD: Norm-guided One-step Denoising Diffusion for Anomaly Detection}.
\newblock
\newblock
\showeprint[arxiv]{2303.08730}~[cs.CV]


\bibitem[Zhang et~al\mbox{.}(2018)]%
        {lpips}
\bibfield{author}{\bibinfo{person}{Richard Zhang}, \bibinfo{person}{Phillip Isola}, \bibinfo{person}{Alexei~A Efros}, \bibinfo{person}{Eli Shechtman}, {and} \bibinfo{person}{Oliver Wang}.} \bibinfo{year}{2018}\natexlab{}.
\newblock \showarticletitle{The unreasonable effectiveness of deep features as a perceptual metric}. In \bibinfo{booktitle}{\emph{Proceedings of the IEEE conference on computer vision and pattern recognition}}. \bibinfo{pages}{586--595}.
\newblock


\bibitem[Zhou and Paffenroth(2017)]%
        {robustae}
\bibfield{author}{\bibinfo{person}{Chong Zhou} {and} \bibinfo{person}{Randy~C Paffenroth}.} \bibinfo{year}{2017}\natexlab{}.
\newblock \showarticletitle{Anomaly detection with robust deep autoencoders}. In \bibinfo{booktitle}{\emph{Proceedings of the 23rd ACM SIGKDD international conference on knowledge discovery and data mining}}. \bibinfo{pages}{665--674}.
\newblock


\bibitem[Zou et~al\mbox{.}(2022)]%
        {visa}
\bibfield{author}{\bibinfo{person}{Yang Zou}, \bibinfo{person}{Jongheon Jeong}, \bibinfo{person}{Latha Pemula}, \bibinfo{person}{Dongqing Zhang}, {and} \bibinfo{person}{Onkar Dabeer}.} \bibinfo{year}{2022}\natexlab{}.
\newblock \showarticletitle{Spot-the-difference self-supervised pre-training for anomaly detection and segmentation}. In \bibinfo{booktitle}{\emph{European Conference on Computer Vision}}. Springer, \bibinfo{pages}{392--408}.
\newblock


\end{thebibliography}


%%% -*-BibTeX-*-
%%% Do NOT edit. File created by BibTeX with style
%%% ACM-Reference-Format-Journals [18-Jan-2012].

\begin{thebibliography}{0}

%%% ====================================================================
%%% NOTE TO THE USER: you can override these defaults by providing
%%% customized versions of any of these macros before the \bibliography
%%% command.  Each of them MUST provide its own final punctuation,
%%% except for \shownote{}, \showDOI{}, and \showURL{}.  The latter two
%%% do not use final punctuation, in order to avoid confusing it with
%%% the Web address.
%%%
%%% To suppress output of a particular field, define its macro to expand
%%% to an empty string, or better, \unskip, like this:
%%%
%%% \newcommand{\showDOI}[1]{\unskip}   % LaTeX syntax
%%%
%%% \def \showDOI #1{\unskip}           % plain TeX syntax
%%%
%%% ====================================================================

\ifx \showCODEN    \undefined \def \showCODEN     #1{\unskip}     \fi
\ifx \showDOI      \undefined \def \showDOI       #1{#1}\fi
\ifx \showISBNx    \undefined \def \showISBNx     #1{\unskip}     \fi
\ifx \showISBNxiii \undefined \def \showISBNxiii  #1{\unskip}     \fi
\ifx \showISSN     \undefined \def \showISSN      #1{\unskip}     \fi
\ifx \showLCCN     \undefined \def \showLCCN      #1{\unskip}     \fi
\ifx \shownote     \undefined \def \shownote      #1{#1}          \fi
\ifx \showarticletitle \undefined \def \showarticletitle #1{#1}   \fi
\ifx \showURL      \undefined \def \showURL       {\relax}        \fi
% The following commands are used for tagged output and should be
% invisible to TeX
\providecommand\bibfield[2]{#2}
\providecommand\bibinfo[2]{#2}
\providecommand\natexlab[1]{#1}
\providecommand\showeprint[2][]{arXiv:#2}

\end{thebibliography}

%%
%% If your work has an appendix, this is the place to put it.
% \appendix

% \section{Research Methods}

% \subsection{Part One}

% Lorem ipsum dolor sit amet, consectetur adipiscing elit. Morbi
% malesuada, quam in pulvinar varius, metus nunc fermentum urna, id
% sollicitudin purus odio sit amet enim. Aliquam ullamcorper eu ipsum
% vel mollis. Curabitur quis dictum nisl. Phasellus vel semper risus, et
% lacinia dolor. Integer ultricies commodo sem nec semper.

% \subsection{Part Two}

% Etiam commodo feugiat nisl pulvinar pellentesque. Etiam auctor sodales
% ligula, non varius nibh pulvinar semper. Suspendisse nec lectus non
% ipsum convallis congue hendrerit vitae sapien. Donec at laoreet
% eros. Vivamus non purus placerat, scelerisque diam eu, cursus
% ante. Etiam aliquam tortor auctor efficitur mattis.

% \section{Online Resources}

% Nam id fermentum dui. Suspendisse sagittis tortor a nulla mollis, in
% pulvinar ex pretium. Sed interdum orci quis metus euismod, et sagittis
% enim maximus. Vestibulum gravida massa ut felis suscipit
% congue. Quisque mattis elit a risus ultrices commodo venenatis eget
% dui. Etiam sagittis eleifend elementum.

% Nam interdum magna at lectus dignissim, ac dignissim lorem
% rhoncus. Maecenas eu arcu ac neque placerat aliquam. Nunc pulvinar
% massa et mattis lacinia.

\end{document}